\def\year{2018}\relax
\documentclass[letterpaper, 10 pt, conference]{ieeeconf}
\IEEEoverridecommandlockouts                              

\usepackage{graphicx}
\usepackage{algorithm}
\usepackage{algorithmic}
\usepackage{color}
\usepackage{subfigure}
\usepackage{amsmath}
\usepackage{amssymb}
\usepackage{amsfonts}
\usepackage{bm}
\usepackage{array}
\usepackage{tikz}
\usepackage{url}
\usepackage{multirow}
\usepackage{color}
\usepackage{colortbl}
\usepackage{adjustbox}
\usepackage{mathtools}

\DeclareMathOperator{\EX}{\mathbb{E}}

\title{\LARGE \bf Accept Synthetic Objects as Real: End-to-End Training of Attentive Deep Visuomotor Policies for Manipulation in Clutter 
}

\author{
Pooya Abolghasemi$^{1}$, Ladislau B{\"o}l{\"o}ni$^{1}$
\thanks{$^{1}$Pooya Abolghasemi and Ladislau B{\"o}l{\"o}ni are with the Dept. of Computer Science,
        University of Central Florida, Orlando, United States
        {\tt\small pooya.abolghasemi@knights.ucf.edu,  lboloni@cs.ucf.edu}}%
}

\begin{document}

\pagestyle{plain}

%



\maketitle
\begin{abstract}
Recent research demonstrated that it is feasible to end-to-end train multi-task deep visuomotor policies for robotic manipulation using variations of learning from demonstration (LfD) and reinforcement learning (RL). In this paper, we extend the capabilities of end-to-end LfD architectures to object manipulation in clutter. We start by introducing a data augmentation procedure called Accept Synthetic Objects as Real (ASOR). Using ASOR we develop two network architectures: implicit attention ASOR-IA and explicit attention ASOR-EA. Both architectures use the same training data (demonstrations in uncluttered environments) as previous approaches. Experimental results show that ASOR-IA and ASOR-EA succeed in a significant fraction of trials in cluttered environments where previous approaches never succeed. In addition, we find that both ASOR-IA and ASOR-EA outperform previous approaches even in uncluttered environments, with ASOR-EA performing better even in clutter compared to the previous best baseline in an uncluttered environment.

\end{abstract}

\section{Introduction}


Recent research demonstrated the feasibility of end-to-end learning of deep visuomotor policies for robot manipulation. Such a policy can be denoted as $\pi(O, T; \phi) \rightarrow a$ where $O$ is an observation involving the vision input of the robot and possibly other sensors such as robot proprioception, $T$ is a specification of the current task and $a$ is a command sent to the robot. The policy is implemented as a neural network with parameters $\phi$. Some approaches propose a learning from demonstration (LfD) model, where the training data consists of demonstrations in the form $D_i = \{T_i, O_{0i}, a_{0i}, \ldots O_{ni}, a_{ni}\}$. An alternative is the use of reinforcement learning (RL), which requires a task-specific reward function $r(O,T)$. The objective of learning is to find an appropriate parameterization $\phi$ of the policy that maximizes some metric of manipulation success. Robot end-to-end learning usually does not take place in a ``big data'' regime, as demonstrations are challenging to collect, reinforcement trials are expensive and dense reward functions difficult to engineer. A series of recent papers had investigated various settings of the problem, ranging from pure LfD over uninitialized networks to pure RL, as well as combinations of LfD and RL. The sparsity of training data puts a special focus on incorporating an appropriate inductive bias into the network structure and learning process. 

In this paper, we consider the case of a robotic arm that was assigned the task to manipulate (push, pick up) a certain object. The task specification has the form $T = \{a_m, (f_{s}, f_{c})\}$ where $a_m$ is the manipulation action while the object to be manipulated is identified by two features: the shape $f_{s}$ and color $f_{c}$. Examples include ``Pick up the white towel'' or ``Push the red bowl". 



One of the challenges which had not yet been consistently solved by deep visuomotor policies is the purposeful manipulation of objects in the presence of random clutter. For the purpose of this paper, we define clutter as objects in the scene that do not need to be manipulated. If a scene contains a bowl and a towel, with the task to pick up the bowl, then the towel is clutter. If the robot needs to push aside the towel to pick up the bowl, then the towel becomes the target object for the push subtask.

In the following, we first discuss why operation in clutter is a particular problem for end-to-end learned policies and then outline our proposed solution. In a typical deep visuomotor policy there is an internal representation bottleneck we will call the {\em primary latent encoding} $\mathbf{z}$ separating the network into a vision component $f_v(\cdot) \rightarrow \mathbf{z}$ and a motor component $f_m(O,T) \rightarrow a$. It is tempting to use an off-the-shelf pretrained network as a vision component such as VGG-19 or ResNet and to keep the size of the encoding small. A low dimensional $\mathbf{z}$ allows us to keep the number of demonstrations and/or reinforcement trials low. For instance, it was found that if $||\mathbf{z}|| \approx 64$, the motor network can be trained from scratch with less than a hundred demonstrations per task, a number which can be further reduced with techniques such as meta-learning. 

Clutter, however, presents a problem to a small, pretrained latent encoding. As the off-the-shelf encoding does not depend on the task specification $T$, the encoding $\mathbf{z} = f_v(O)$ will need to represent the entire scene, leaving it to the motor network to sort out what is important. If $||\mathbf{z}||$ is small, there is simply no space for an accurate representation for many objects in the scene. The quality of representation for the target object will suffer, impacting the success rate of the policy. 

There are several ways this problem can be solved. We can increase the size of $||\mathbf{z}||$ - but this requires a corresponding, usually higher-than-linear increase in the training data. For instance, we would need to provide a large number of demonstrations done in various clutter scenarios. Another possibility is to use a higher dimensional encoding $\mathbf{z}$, but enforce on it a pre-defined object-oriented representation~\cite{jiang2019language}. A benefit of this approach is that it is more explainable. However, to some degree, it backs off from the end-to-end learned paradigm towards more engineered approaches.

The approach we take in this paper retains the fully end-to-end trained model. As a first step, we require the visual component of the network to create a primary latent encoding that depends on the current manipulation task $\mathbf{z} = f_v(O, T)$ - this can be seen as a way to reserve a larger part of the representation to the target object. Nevertheless, this does not entirely eliminate the need for a variety of demonstrations in conditions of clutter. Our approach is based on the observation that such extra demonstrations convey new visual data, but very little new motion information. Human demonstrators already have a mental mechanism for ignoring clutter -- thus they will usually deliver the same robot trajectory whether clutter is present or not. This leads us to the idea that it should be possible to train a visuomotor policy that performs under clutter conditions without requiring {\em any} demonstration in clutter. 


The contributions of this paper are as follows. (1) We introduce a data augmentation technique and associated training regime called Accept Synthetic Objects as Real which allows for the generation of training data suitable for training manipulation in the clutter from demonstrations collected in scenarios without clutter. (2) We introduce a network architecture ASOR-Implicit Attention that trains a visual representation that implicitly encodes an attention feature on the target object in the primary latent encoding and its associated motor component. (3) We introduce a network architecture ASOR-Explicit Attention which trains a visual representation that explicitly focuses on the target object in the form of spatial attention, as well as its associated motor component. Both approaches are trained with demonstrations generated by ASOR. (4) We evaluate the two approaches on a series of manipulation experiments and compare them with previous approaches trained on the same data. The results show that ASOR-IA and ASOR-EA succeed in a significant fraction of trials in cluttered environments where previous approaches never succeed. In addition, we find both ASOR-IA and ASOR-EA outperform previous approaches even in uncluttered environments, ASOR-EA performs better than ASOR-IA, and ASOR-EA performs better even in clutter compared to the previous approaches in an uncluttered environment.

\section{Related Work}

The canonical approach for robot manipulation involves decomposing the task into several well-defined components such as state estimation, planning, and low-level control, which often have further subcomponents such as a model of environment's dynamics in the planner. Recent advances in deep learning led to solutions where some of these components are implemented as neural networks. Examples include object pose estimation~\cite{rahmatizadeh2016virtual}, robot pose estimation~\cite{zuo2019craves}. In some cases, more than one component is implemented with neural networks, such as the inference of grasp and suction affordances and recognizing novel objects \cite{zeng2018robotic}.

Taking this trend to the extreme are end-to-end learned deep visuomotor policies where all or almost all the canonical pipeline is implemented as a {\em single} neural network, transforming in one step a visual input into the control signal~\cite{levine2016end,rajeswaran2017learning}. The challenge, of course, is that we need to learn a large, opaque neural network, losing the advantages of decomposition and module testing. 

The two major learning models of end-to-end policies are reinforcement learning RL and learning from demonstration LfD. For RL, the task is specified through a reward function and requires the robot to interact with the environment in order to collect training data. An approach to avoid the requirement of reward engineering by periodic querying of the user is shown in~\cite{singh2019end}. Although some projects collected a large number of reinforcement trials~\cite{levine2018learning}, for most projects the goal is to reduce the number of physical trials required. This can be done by training in simulation with testing done in the same simulated environment~\cite{devin2017learning, rajeswaran2017learning} or the learned behavior will be transferred to a real robot~\cite{zhu2018reinforcement}. Another alternative is to use LfD to bootstrap RL and overcome exploration challenges~\cite{rajeswaran2017learning, zhu2018reinforcement}.

LfD has the advantage that it does not require an engineered reward function, the tasks being specified by demonstrations performed by the user. In a very primitive setting, using only a behavioral cloning/imitation loss, LfD would be limited to reproducing the trajectories of the user, with fast divergence whenever a new situation is encountered. In practice, there are many techniques that can learn well-generalizing policies from demonstrations by combining the imitation loss with other losses, regularization techniques and other ways to introduce inductive bias. Just like in the case of RL, it is desirable to reduce the number of demonstrations required to learn a new task. One possible approach for this is meta-learning: by utilizing demonstrations from a variety of tasks, under some conditions it is possible to learn a new task from a single demonstration~\cite{yu2018one}. 

One of the features of our model is the use of attention in the representation of the visual input. Attention is a technique which both humans and AI use to process a large amount of data in real-time. Attention can caption images~\cite{xu2015show, you2016image}, answer questions based on videos~\cite{mazaheri2017video, yang2016stacked}, help robots navigate through crowds~\cite{vemula2018social}. In ~\cite{devin2017deep} the authors use a set of demonstrations to learn the attention among region proposals from a pre-trained network on the MSCOCO dataset~\cite{lin2014microsoft} to create an object-centric representation to be used by an RL-learned robot manipulation policy. This can be contrasted to our approach where the attention is obtained in a single step, it does not rely on external training data, and its focus object is specified from a natural language description of the task. 



\section{Accept Synthetic Objects as Real}

Our objective is to teach a robot arm to manipulate objects of different types under conditions of clutter. We will perform this by training a visuomotor policy that takes as input a video stream of the scene and generates commands to the robot. To be able to make a one-to-one comparison to previous approaches, we will reuse one of the existing datasets for which both training data and code for previous approaches is publicly available \cite{Abolghasemi_2019_CVPR}. This scenario includes training data for picking up 8 different object types and pushing to place 5 different object types. The objects are distinguished by shape and color, and have different degrees of rigidity including a rigid plastic bowl, a foam dumbbell, a piece of bubble wrap and a cotton towel. 


\begin{figure}[ht]
    \centering
    \includegraphics[width=\columnwidth]{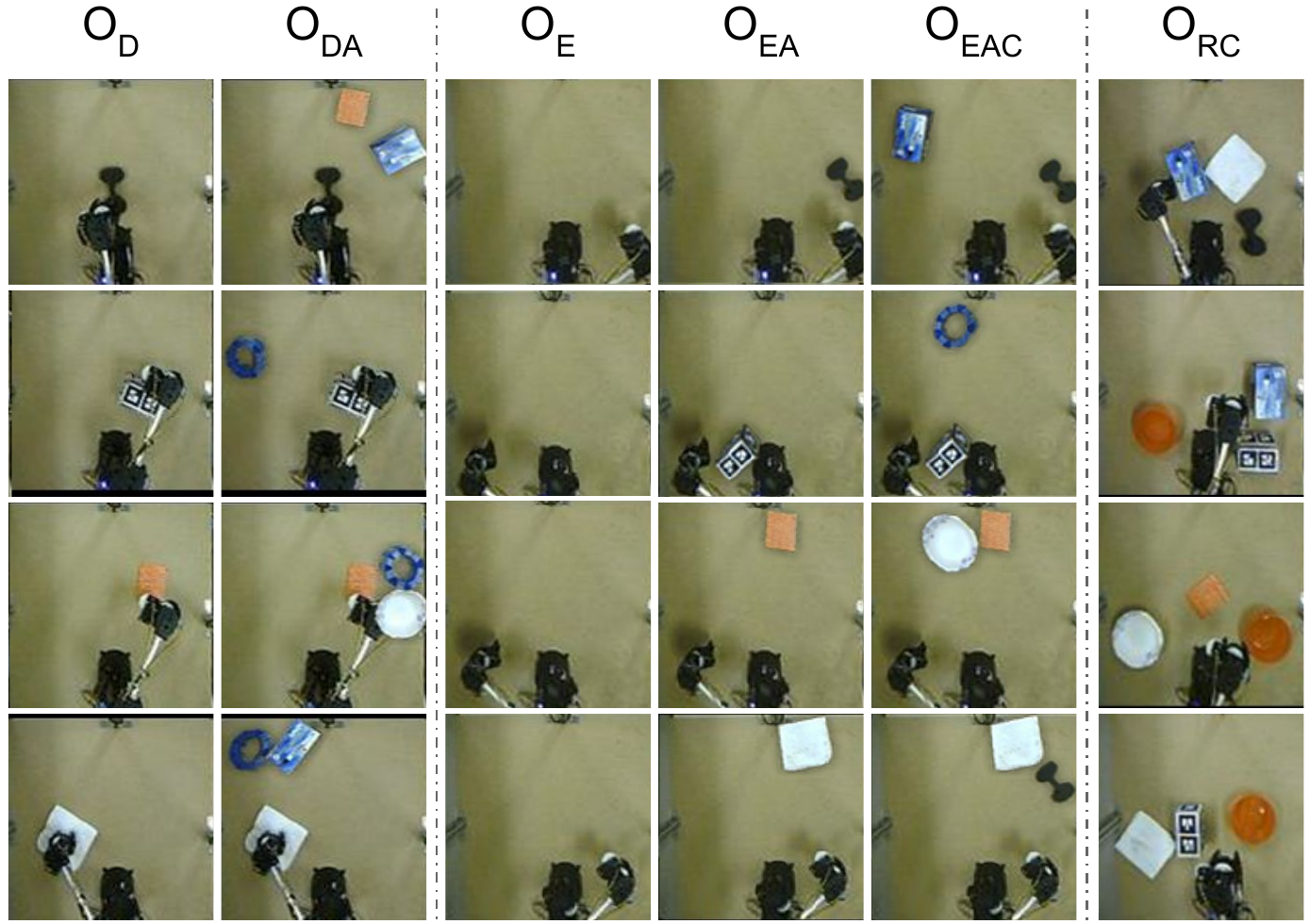}
    \caption{Faking clutter and accepting it as real.
    \textbf{$O_D$:} Demonstrations captured in an uncluttered environment~\cite{Abolghasemi_2019_CVPR}.
    \textbf{$O_{DA}$:} $O_D$ images augmented by synthetic clutter. 
    \textbf{$O_E$:} Empty images contain only the robot.
    \textbf{$O_{EA}$:} $O_E$ images augmented by a synthetic target object.
    \textbf{$O_{EAC}$:} $O_{EA}$ images augmented by synthetic clutter.
    \textbf{$O_{RC}$:} Real clutter(random objects in random positions).
}
    \label{fig:data_aug}
    \vspace{-5mm}
\end{figure}

The training data contains manipulation demonstrations done without the presence of clutter, with the only objects visible in the scene being the object to be manipulated and the robot arm itself (Figure~\ref{fig:data_aug} Demonstration Observation column or $O_D$). 

As we discussed in the introduction, collecting training data in the presence of clutter is not an ideal way to use human effort, as humans will simply ignore the clutter. A better approach would augment the training data by adding clutter objects after the fact, using a visual manipulation of the images. This is shown in Figure~\ref{fig:data_aug} Augmented Demonstration Observation or $O_\mathit{DA}$.

Unfortunately, this approach, by itself, does not work. While copy-pasted synthetic objects may look natural to the human eye, they have low-level artifacts that allow the vision network to recognize them. As in these images, the target object is always real, and clutter objects are synthetic, the network learns to ignore clutter based on its synthetic nature. Such policies are not able to ignore clutter when tested with real objects. Attempts to improve our image processing capabilities to the degree that the artificially created clutter would be indistinguishable from the real one would be fighting against the training of the vision system which is motivated to distinguish them. 

As our object is not to generate more convincing images but to train the robot, we propose an approach that intervenes in the end-to-end training loss of the robot. Instead of improving the synthetic objects to be indistinguishable from real objects, we train the vision system to Accept Synthetic Objects as Real. This technique requires both specific type of generated training images as well as training regime and losses relying on them. Columns 3-5 in Figure~\ref{fig:data_aug} show the new type of training images generated. We started by recording a small set of images with only the robot arm in random positions and no other object (Empty Observations or $O_E$). Then we created images with one synthetic target object (Augmented Empty Observations or $O_\mathit{EA}$) and with a synthetic target object and several synthetic clutter objects (Clutter Augmented Observations or $O_\mathit{EAC}$). Note that there is no visual difference between the target object and the clutter objects: all of them are synthetic, the only feature of the target object is that it is referred to in the task specification. We can generate an arbitrary amount of training data through this technique. However, this training data is useful only for training the vision component, as it does not contain purposeful robot arm motion. We have another type of input images which we named Real Clutter Observations or $O_\mathit{RC}$ which contains random real objects in random positions to give the model more generalization power.

With this set of images we use a training regime where the vision system does not distinguish between synthetic and real objects: it can accept the synthetic target object as real one in columns $O_{EA}$ and $O_{EAC}$, and it can remove the clutter objects from $O_{EAC}$ to create a generated scene as in $O_{EA}$. 

In the following two sections we describe two different network architectures for visuomotor policies that take advantage of the ASOR model. We start by recognizing that any model where the primary latent encoding $\mathbf{z}$ prioritizes the target object specified by the task specification can be seen as attention based. This can be an {\em explicit attention} where the network actually creates a heatmap style 2D overlay for the visual field, a model used in ASOR-EA discussed in Section~\ref{sec:explicit}. Alternatively, we can use ASOR in a network where such an explicit model does not exist -- an {\em implicit attention} model such as the ASOR-IE discussed in Section~\ref{sec:implicit}.
\section{ASOR With Implicit Attention}
\label{sec:implicit}

\begin{figure*}[ht]
    \centering
    \vspace{1.5mm}
    \includegraphics[width=0.9\textwidth]{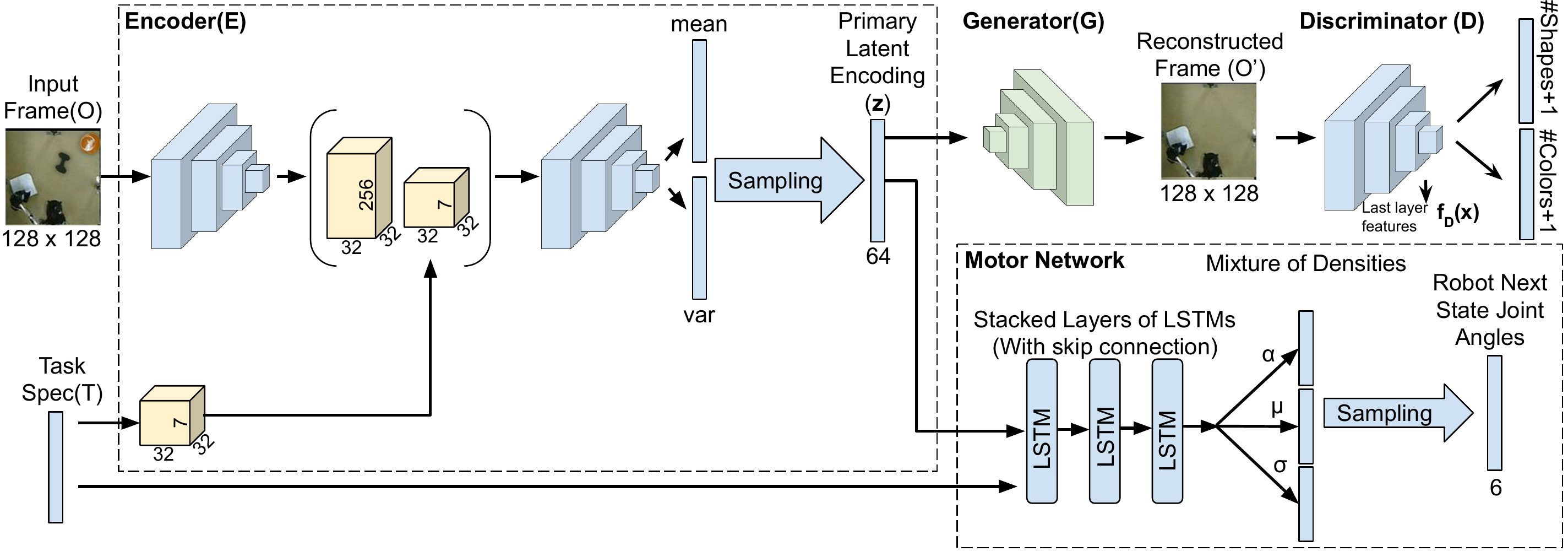}
    \caption{The ASOR-IA network architecture. In each time-step the \textbf{Encoder(E)} receives the task specification and an image($O$) as input. Features extracted from the task specification will be concatenated to feature maps extracted from the image somewhere in the middle of the CNN layers. The \textbf{Generator(G)} is trained to reconstruct the image given the primary latent encoding $\mathbf{z}$. The reconstructed image should only contain the target object and ignore all the unnecessary information. $\mathbf{z}$ and the task specification is then passed to the motor controller to generate the next robot joint angles.}
    \label{fig:arch_simple}
    \vspace{-5mm}
\end{figure*}

The ASOR-IA network architecture, shown in Figure~\ref{fig:arch_simple} uses a VAE-GAN based visual network evolved from the one introduced in~\cite{rahmatizadeh2017vision} featuring an encoder E, generator G, and discriminator D. 



Next, we discuss the discriminator's loss function $\mathcal{L}_{D}$, and the generator's loss function $\mathcal{L}_{G}$. All the parameters in the Discriminator have been optimized to minimize $\mathcal{L}_{D}$, and parameters of the Encoder, Generator and the motor network are optimized by the loss value $\mathcal{L}_{G}$.


The discriminator classifies the input frames based on target object's shape($s$) and color($c$), it also will classify fake frames in a separate class~\cite{salimans2016improved}. The discriminator's outputs are of lengths $|s| + 1$ and $|c| + 1$. $|s|$ and $|c|$ are respectively the number of unique shapes and colors and the ``$+1$'' represents the ``fake'' class. If the discriminator $D$ receives a real frame from $\mathcal{O}$, it needs to classify the frame correctly in the corresponding object's shape and object's color class provided in the task specification (`Push the \underline{red} \underline{bowl}/pick-up the \underline{white} \underline{towel}`):
\begin{align}
    \label{eq:dis_real}
    \mathcal{L}_\mathit{real} =& - \EX_{\mathcal{O}, s \sim p_\mathit{data}} [\log{(P_\mathit{D}(s \big\vert \mathcal{O}))}]\nonumber\\
    &- \EX_{\mathcal{O}, c \sim p_\mathit{data}} [\log{(P_\mathit{D}(c \big \vert \mathcal{O}))}]
\end{align}
\noindent where $\mathcal{O} \in \{ \mathcal{O_D}, \mathcal{O}_{EA}\}$ and $P_{\mathit{D}}$ is the class probabilities produced by the discriminator for object's shape and color. Similarly, if $D$ receives fake frames generated by the generator, it should classify them as fake:

\begin{align}
    \mathcal{L}_\mathit{fake} =& - \EX_{\mathcal{O}' \sim G} [\log{(P_\mathit{D}(|s|+1 \big \vert \mathcal{O}'))}]\nonumber\\
            &- \EX_{\mathcal{O}' \sim G} [\log{( P_\mathit{D}(|c|+1 \big \vert \mathcal{O}'))}]
\end{align}
\noindent where $\mathcal{O}'$ is the generator's reconstruction of any of the 5 input types. Finally, if $D$ receives fake frames, generated by $G$ with the latent representation $\mathbf{z} \sim \mathcal{N}(0, 1)$:
\noindent \begin{align}
    \mathcal{L}_\mathit{noise}\!=\!
    & - \EX_{z \sim \mathit{noise}} [\log{( P_\mathit{D}(|s|+1 \big \vert G(z)))}] \nonumber\\
    &- \EX_{z \sim \mathit{noise}} [\log{(P_{D}(|c|\!+\!1 \big \vert G(z)))}].
\end{align}

The overall loss of the discriminator is thus $\mathcal{L}_{D} = \mathcal{L}_\mathit{real} + \mathcal{L}_\mathit{fake} + \mathcal{L}_\mathit{noise}$.

The Generator(G) must reconstruct a real looking frame($O'$) that only contains the object described in the input sentence. In fact, $G$ tries not only to look real, but also presents the correct object in its output.
\begin{align}
    \mathcal{L}_\mathit{GD} =& - \EX_{\mathcal{O}', s \sim p_G} [\log{P_\mathit{D}(s \big \vert \mathcal{O}')}]\nonumber\\
    &- \EX_{\mathcal{O}', c \sim p_G} [\log{P_\mathit{D}(c \big \vert \mathcal{O}')}].
\end{align}

The generator needs to learn to remove the irrelevant objects present in the scene. The discriminator as described in equation~\ref{eq:dis_real}, only receives single object images. As a result, it should be able to easily classify multi object images as fake. Moreover, to enforce such behavior and to stabilize the training process we use feature matching~\cite{salimans2016improved} technique and the reconstruction error. Using these two terms in our loss function, we will force the network to reconstruct images of type $O_{DA}$ and $O_{EAC}$ as close as possible to the corresponding images of type $O_D$ and $O_{EA}$.
\begin{align}
\mathcal{L}_\mathit{rec} = ||O'_{D} - O_D||^2 &+ ||O'_{DA} - O_D||^2 \nonumber\\
                         + ||O'_{EA} - O_{EA}||^2 &+ ||O'_{EAC} - O_{EA}||^2.    
\end{align}
 \begin{align}
    \mathcal{L}_\mathit{fea} = ||f_{D}(O'_D) - f_{D}(O_D)||^2 &+ ||f_{D}(O'_{DA}) - f_{D}(O_D)||^2 \nonumber\\
    + ||f_{D}(O'_{EA}) - f_{D}(O_{EA})||^2 &+ ||f_{D}(O'_{EAC}) - f_{D}(O_{EA})||^2
\end{align}
\noindent where $f_D(x)$ is features extracted by the discriminator in the last layer for input $x$. To regularize the primary latent encoding ($\mathbf{z}$), we minimize the KL-divergence between $\mathbf{z}$ and $\mathcal{N}(0,1)$:

\begin{align}
    \mathcal{L}_\mathit{prior} &= D_\mathrm{KL}(E(O, T)~||~\mathcal{N}(0, 1)).
\end{align}

Finally, we write the generator's loss as $\mathcal{L}_{G} = \mathcal{L}_\mathit{DG} + \mathcal{L}_\mathit{rec} + \mathcal{L}_\mathit{prior} + \mathcal{L}_\mathit{motor}$, where the $\mathcal{L}_\mathit{motor}$ is calculated based on the behaviour cloning loss similar to~\cite{rahmatizadeh2017vision}.

\section{ASOR with Explicit Attention}
\label{sec:explicit}

\begin{figure*}[ht]
    \centering
    \vspace{1.5mm}
    \includegraphics[width=0.9\textwidth]{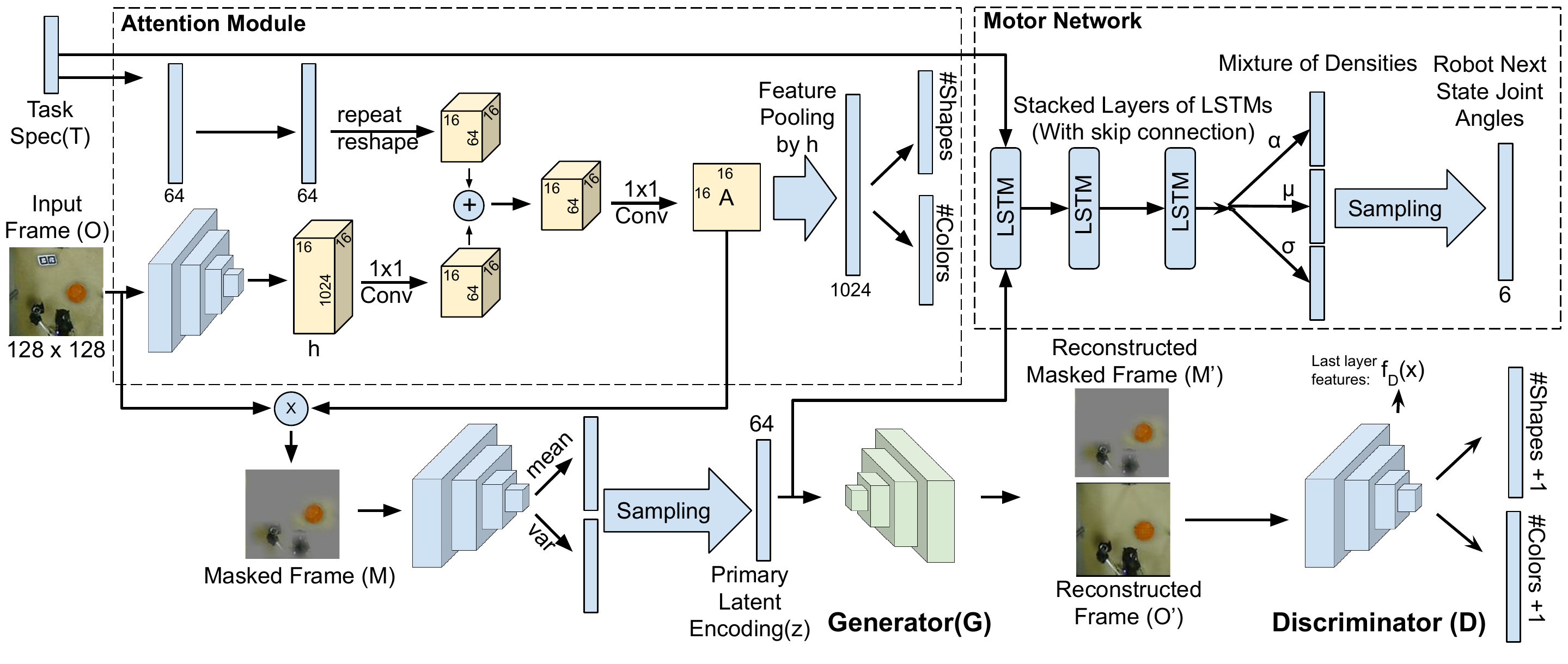}
    \caption{The ASOR-EA network architecture. The task specification and the image capture are the inputs to the attention network. The attention network generates the attention map and should be able to classify the input image based on the features pooled from feature maps pooled from the last convolution layer of the attention map. The VAE-GAN will receive a masked image of the original input by the attention map and should be able to reconstruct the input image without the clutter objects.}
    \label{fig:vae_att}
    \vspace{-5mm}
\end{figure*}

The second architecture we discuss, ASOR-EA generates explicit attention maps for the input frames. The network, shown in Figure~\ref{fig:vae_att} has 5 primary components: encoder, generator, discriminator, attention and the motor network. First, the attention module extracts an attention mask to cover the parts of the input frame that are not relevant to the current task (the clutter). The masked frame $M$ is the result of pixel-wise multiplication of the attention map $A$ and the input frame $O$. Given the masked frame $M$, the encoder-generator pair will try to extract a set of features descriptive enough to create reconstructed versions of the original frame $O'$ and of the masked frame $M'$. The reconstruction of the masked frame $M'$ is necessary to ensure that the information regarding the attention map is preserved in the latent encoding to be used later by the motor controller. $O'$ is fed to the discriminator network.

\begin{figure}
    \centering
    \includegraphics[width=0.75\columnwidth]{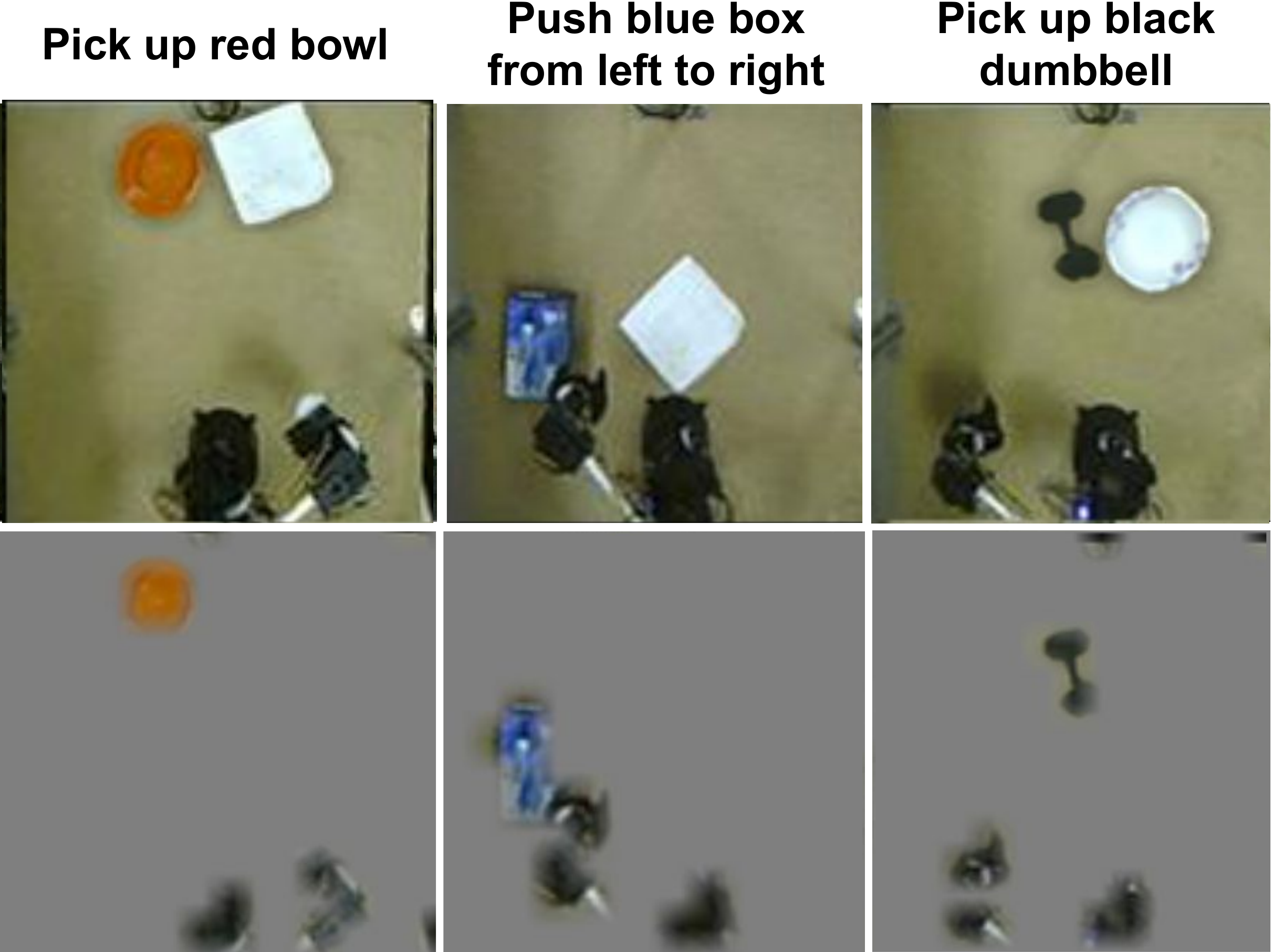}
    \caption{Examples of the operation of the attention module in ASOR-EA. The first row is the input image $O$ and the second row is the masked image $M$ for the task specification on the top of the column.}
    \label{fig:TFA-example-last}
    \vspace{-5mm}
\end{figure}

The attention network combines features from the task specification and the input frame to extract the attention map. Several convolutional layers augmented by batch normalization are used to obtain $K$ spatial visual feature vectors. These convolution layers will provide spatial visual features with size $\phi_{f} \in \mathcal{R}^{K \times d_\phi}$, where $d_\phi$ is the number of features for region $k_i$. The task specification encodes the target object as two concatenated one-hot vectors describing the shape and color of the object respectively. Several fully-connected layers are used to transform the task encoding to shape $u \in \mathcal{R}^{d_\psi}$. In order to combine the visual and task features repeated the $u$ vector $k$ times and combined it with the visual features using a technique similar to~\cite{yang2016stacked,mazaheri2017video}:
\begin{equation}
    \psi = \tanh(\phi_f \times W_f \oplus u),
\end{equation}
\noindent where $W_f \in \mathcal{R}^{d_\phi \times d_\psi}$ is a mapping matrix, $\oplus$ is element-wise summation. The combined feature vector is of size $\mathcal{R}^{k \times d_\psi}$.
To compute the final attention map we calculate
\begin{align}
    &p = \mathrm{sigmoid}(\psi \times W_p),\\
    &p_\textit{TFA} = \mathrm{ReLU}(p - t),
\end{align}
\noindent where $W_p \in \mathcal{R}^{d_{\psi} \times 1}$ represents a fully-connected layer, and $t$ is a hyper-parameter. $t$ can be a constant or it can be a statistical metric such as a running average over $p$. Here we set $t$ to the constant value of $0.5$. $\mathrm{ReLU}$ and the hyper-parameter $t$ are needed to force the network to completely hide the information from the irrelevant regions by assigning a score of exactly $0$ to most regions. Without these, the attention map will keep small values in most regions and the encoder can easily reconstruct the original image. The final $p_\textit{TFA} \in [0,1]^{k}$ contains attention scores of all $k$ regions.

Similar to the teacher network in~\cite{Abolghasemi_2019_CVPR} the network should be able to reconstruct the task specification from the pooled spatial features weighted by the attention map, $p_\textit{TFA}$. Weighted pooled features are defined as $p_f = \sum_{i \in k}{p_{\textit{TFA}_i} {\phi_f}_i }$ where $p_f \in \mathcal{R}^{d_{\phi}}$ and are passed to two multi-layer perceptrons (MLP) $\hat{s} =\tau_{1}(p_f)$, $\hat{c} =\tau_{2}(p_f)$. These MLP layers try to classify the pooled features based on target object's shape and color. In other words, the network is trying to focus the attention on regions which provide better information to reconstruct the task specification and an image which only contains the target object. 

We use the L1 norm of the $p_{\textit{TFA}}$ and softmax cross entropy between the $f_s$ and $f_c$ from the task specification and $(\hat{s}, \hat{c})$ to calculate $\mathcal{L}_{A}$. 


The masked frame $M$ is generated using the attention map, $M = O \otimes p_{\textit{TFA}}$ where $\otimes$ is element-wise multiplication. Examples of input frames and the masked frames using the computed attention $p_{\textit{TFA}}$ are shown in Figure~\ref{fig:TFA-example-last}.

The encoder $E$ receives $M$ as input, $[\mu_{\mathbf{z}} | \sigma_{\mathbf{z}}] = E(M)$ where $\mu_{z}$, $\sigma_{z} \in \mathcal{R}^{d_z}$, and $d_z$ is the length of the primary latent encoding $\mathbf{z}$. $E$ is a multi-layer convolutional neural network with a $2d_z$ dimensional vector output split into $\mu_z$ and $\sigma_z$. We assume that $\mathbf{z} \sim \mathcal{N}(\mu_{z},\sigma_{z})$. The generator receives $\mathbf{z}$ and reconstructs the input frame $O$ and the masked frame $M$. Note that in this case information regarding the target object and the text encoding is not passed to the encoder. The encoder should rely solely on the masked frame to extract the required information for $G$ to reconstruct the original frame and the masked frame. In addition to the $\mathcal{L}_{A}$, the reconstruction loss between different types of input images will also force the attention to focus on parts of the image related to the task.

The loss functions of the discriminator, generator and motor network are similar to the one for ASOR-IA. The only difference being that $\mathcal{L}_{A}$ and the reconstruction loss for the masked frame $M'$ is added to the generator's loss $\mathcal{L}_{G}$.
 

\section{Experiments}
\label{sec:clutter_experiments}

\begin{figure}
    \centering
    \vspace{1.5mm}
    \includegraphics[width=\columnwidth]{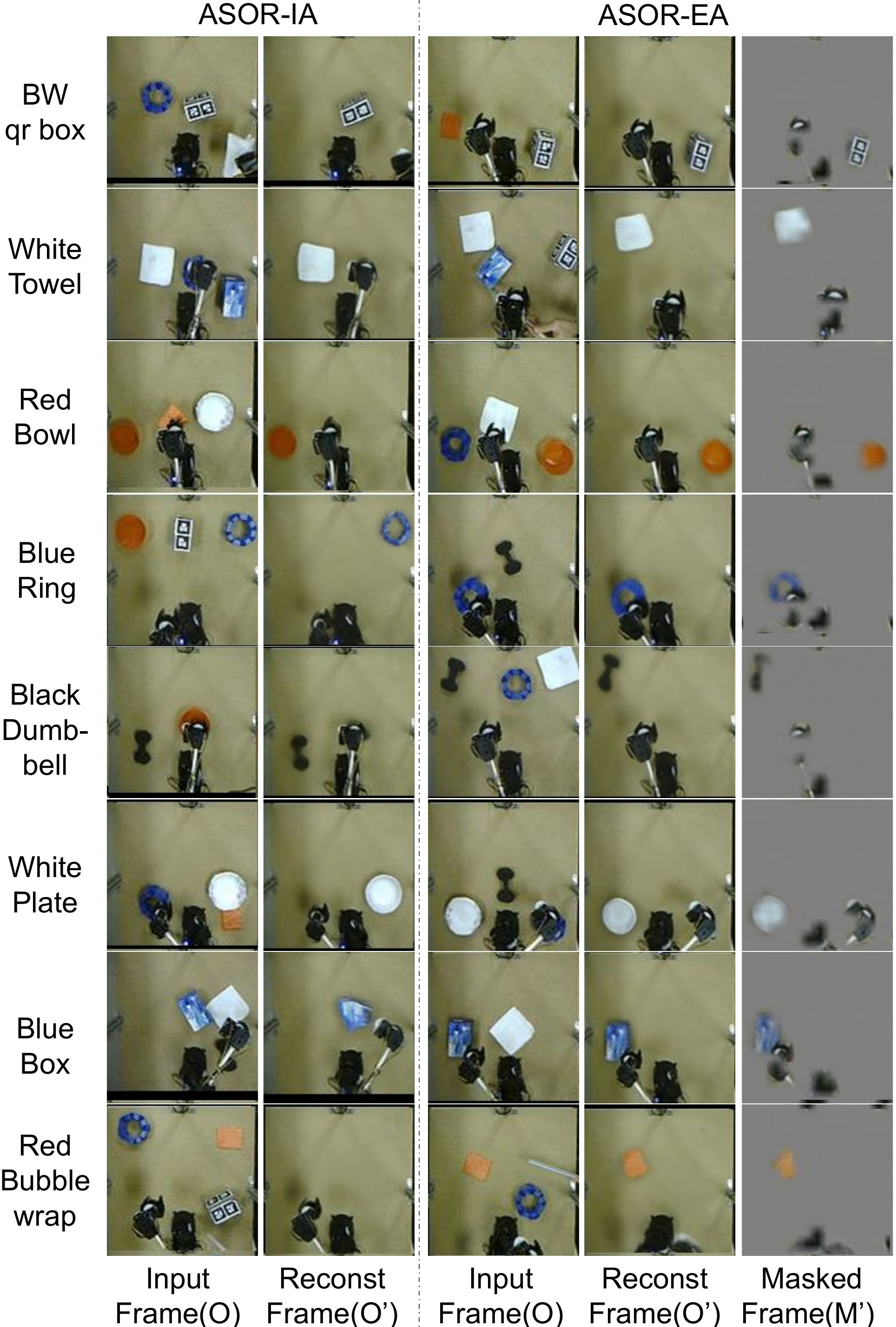}
    \caption{Each row illustrates ASOR-IA and ASOR-EA reconstructions for a specific
    object. First column is the input frame($O$) to ASOR-IA and the second column is
    it's output($O'$). The third column is the input frame($O$) to ASOR-EA and the next two columns are the generator's reconstructions $O'$ and $M'$.}
    \label{fig:results}
    \vspace{-5mm}
\end{figure}

Comparing robot control approaches is a general challenge due to difficult to reproduce environmental settings. It is to be hoped that in future years with the spread of inexpensive, replicable platforms such as REPLAB~\cite{yang2019replab} and PyRobot~\cite{murali2019pyrobot}, such one to one comparisons will be easier to make. For this paper, we will test the proposed approach over 10 tasks for which training data is publicly available~\cite{Abolghasemi_2019_CVPR}. The four algorithms we compare are the one from~\cite{rahmatizadeh2017vision} which does not use an attention model, the algorithm from~\cite{Abolghasemi_2019_CVPR} which uses a language induced attention model called TFA, ASOR-IA as described in Section~\ref{sec:implicit} and ASOR-EA as described in Section~\ref{sec:explicit}. All algorithms were trained with the same demonstrations. 

\definecolor{mygray}{RGB}{225,225,225}
\begin{table}[ht]
    \vspace{1.5mm}
    \centering
    \setlength{\tabcolsep}{5pt}
    \caption{Success percentages: no clutter / with clutter}
    \label{tab:last_results}
    \begin{tabular}{|p{0.12cm}p{1.7cm}p{1.1cm}p{1.1cm}p{1.2cm}p{1.25cm}|}
    \hline
    \multicolumn{2}{|c}{Task spec.} & {\scriptsize w/o att. \cite{rahmatizadeh2017vision}} & {\scriptsize w/ att. \cite{Abolghasemi_2019_CVPR}} & ASOR-IA & ASOR-EA  \\
    \hline    
    \multirow{7}{*}{\rotatebox[origin=c]{-90}{Pick up}} & Red Bowl & 70 / 0 & 80 / 0  & 80 / 60 & {\bf 90} / \bf{80} \\
    & White Towel & 50 / 0  & 80 / 0 & {\bf 100} / 70 & {\bf 100} / {\bf 100} \\
    & Blue Ring & 30 / 0 & 60 / 0 & 60 / 40 & {\bf 80} / {\bf 80} \\
    & {\scriptsize Black Dumbbell} & 40 / 0  & 50 / 0 & 40 / 50 & {\bf 60} /{\bf 70} \\
    & White Plate & 60 / 0 & 80 / 0 & 80 / 60 & {\bf 100} / {\bf 90} \\
    & {\scriptsize Red Bubblewrap} & 10 / 0 & \bf{40} / 0 & 30 / {\bf 20} & {\bf 40} / {\bf 20} \\
    & \cellcolor{mygray} All pick up &\cellcolor{mygray} 43.3 / 0.0 & \cellcolor{mygray}65.0 / 0.0 & \cellcolor{mygray}65.0 / 50.0 &\cellcolor{mygray} {\bf 78.3} / {\bf 70.0} \\
    \multirow{5}{*}{\rotatebox[origin=c]{-90}{Push to right}} & Red Bowl & 80 / 0 & {\bf 100} / 0 & 80 / 80 & {\bf 100} / {\bf 100} \\
    & White Plate & 60 / 0 & 60 / 0 & {\bf 70} / 50 & 60 / {\bf 70} \\
    & Blue Box & 10 / 0 & 30 / 0 & 60 / 50 & {\bf 90} / {\bf 90}\\
    & BW QR-box & 20 / 0 & 60 / 0 & 60 / 40 & {\bf 80} / {\bf 70} \\
    & \cellcolor{mygray} All push tasks & \cellcolor{mygray}42.5 / 0 & \cellcolor{mygray}62.5 / 0 &\cellcolor{mygray} 67.5 / 57.0 & \cellcolor{mygray}{\bf 82.5} / {\bf 82.5} \\
    \hline
    \rowcolor{mygray} \multicolumn{2}{|c}{Overall} & 43.0 / 0.0 & 64.0 / 0.0 & 66.0 / 53.0 & \bf{80.0} / \bf{75.0} \\
    \hline
    \end{tabular}
    \vspace{-5mm}
\end{table}

The task set was composed of 6 tasks of picking up objects (a red bowl, a white towel, a blue ring, a black foam dumbbell, a white plate and a red bubblewrap) and 4 tasks for pushing objects (a red bowl, a white plate, a blue box and a black-and-white QR patterned box). Figure~\ref{fig:results}  illustrates ASOR-IA and ASOR-EA reconstructions for each of these objects. For each algorithm, we repeated the experiments 10 times under two types of conditions: in an uncluttered scene with only the target object and the robot visible and in a cluttered scene where 2-3 other objects were also present. A trial was counted as a success if the manipulation was completed in 2 minutes. 

Table~\ref{tab:last_results} shows the experimental results as percentages of the successes for the no-clutter / clutter case, with the best results for each task setting highlighted in {\bf bold}\footnote{Checkout our YouTube video \url{https://youtu.be/GchuLQhG3ug}}. 

As expected, we find that for almost all trials, the results were worse in clutter. Neither of the previous approaches managed to complete any trial successfully in the presence of clutter. Note that the architecture proposed in~\cite{Abolghasemi_2019_CVPR} is trained with task-focused attention and is able to ignore physical and visual disturbances like a human hand and generally disturbances that are not present in the training set but clutter in our experiments contains objects involved in other demonstrations. ASOR-IA and ASOR-EA outperformed the previous approaches in the no-clutter setting, with ASOR-EA being the better of the two. More importantly, they managed to complete a significant portion of the tasks in clutter as well - in fact, ASOR-EA outperforms in clutter the previous best algorithm operating in an uncluttered setting\footnote{Code is available at \url{https://github.com/pouyaAB/Accept_Synthetic_Objects_as_Real}}.

\section{Conclusion and Future Work}
\label{sec:conclusion}

In this paper, we proposed a data augmentation technique (Accept Synthetic Objects as Real) and two network models that take advantage of it to train end-to-end robot controllers which operate in the presence of clutter. Immediate extensions include more complex manipulation tasks with multiple objects. We believe, however, that the overall idea of co-training the network to accept simplified representations as real has applications in many other areas where robust, task-dependent representations need to be learned from limited training data.





\noindent{\bf Acknowledgments:} This work had been supported in part by the National Science Foundation under grant numbers IIS-1409823. 
\bibliographystyle{IEEEtran}
\bibliography{main}
\clearpage
\end{document}